\documentclass{article} 
\usepackage{iclr2026_conference,times}

\usepackage{amsmath,amsfonts,bm}

\def\eqref#1{equation~\ref{#1}}

\def\1{\bm{1}}

\DeclareMathAlphabet{\mathsfit}{\encodingdefault}{\sfdefault}{m}{sl}
\SetMathAlphabet{\mathsfit}{bold}{\encodingdefault}{\sfdefault}{bx}{n}

\usepackage{hyperref}
\usepackage{url}
\usepackage{cite}
\usepackage{algorithm,algorithmic}
\usepackage{amsmath,amssymb,amsfonts}
\usepackage{algorithmic}
\usepackage{graphicx}
\usepackage{textcomp}
\usepackage{xcolor}
\usepackage{xspace}
\usepackage{caption}
\usepackage{textcomp}
\usepackage{hyperref}
\usepackage{pifont}
\usepackage{caption} 
\usepackage{booktabs}    
\usepackage{multirow}   
\usepackage{amsmath}     
\usepackage{makecell}
\usepackage{natbib} 
\usepackage{hyperref}
\usepackage{graphicx}       
\usepackage{subcaption}     
\usepackage{calc}    
\usepackage{tabularx}
\usepackage{booktabs} 
\usepackage{wrapfig} 
\usepackage{float} 
\usepackage{capt-of} 
\usepackage{tabularx} 
\usepackage{tabularx}
\usepackage{booktabs}
\usepackage{multirow}
\usepackage{makecell}
\usepackage{array}
\newcolumntype{Y}{>{\centering\arraybackslash}X}
\title{Towards Flash Thinking via Decoupled Advantage Policy Optimization}

\author{
  Zezhong Tan\textsuperscript{1,2} \quad
  Hang Gao\textsuperscript{1} \quad
  Xinhong Ma\textsuperscript{1} \quad
  Feng Zhang\textsuperscript{1} \quad
  Ziqiang Dong\textsuperscript{1}\thanks{Corresponding author.}
  \vspace{0.5em} \\ 
  \textsuperscript{1}Alibaba Inc., \textsuperscript{2}Peking University
}

\newcommand{\myparagraph}[1]{\textbf{#1.}\quad}
\def\BibTeX{{\rm B\kern-.05em{\sc i\kern-.025em b}\kern-.08em
    T\kern-.1667em\lower.7ex\hbox{E}\kern-.125emX}}

\newcommand{\eg}{{\emph{e.g.}}\xspace}
\newcommand{\ie}{{\emph{i.e.}}\xspace}

\newcommand{\lrm}{{\textsc{LRMs}}\xspace}
\newcommand{\dec}{{\textsc{DEPO}}\xspace}
\newcommand{\grpo}{{\textsc{GRPO}}\xspace}
\newcommand{\grm}{{\textsc{GRM}}\xspace}

\newcommand{\Overthink}{{\textsc{OverThink}}\xspace}
\newcommand{\distill}{{\textsc{DeepSeek-Distill-Qwen-7B}}\xspace}
\newcommand{\distills}{{\textsc{DeepSeek-Distill-Qwen-1.5B}}\xspace}
\newcolumntype{C}{>{\centering\arraybackslash}X} 
\iclrfinalcopy 
\makeatletter
\renewcommand{\@makefnmark}{\hbox{\@textsuperscript{\normalfont\@thefnmark}}}
\makeatother
\begin{document}

\maketitle
\begin{abstract}
    Recent Large Reasoning Models (\lrm) have achieved remarkable performance in solving complex problems via supervised fine-tuning (SFT) and reinforcement learning (RL). Although existing RL algorithms significantly enhance model accuracy, they still suffer from excessively lengthy responses and overthinking issues, resulting in increased inference latency and computational consumption, especially for simple tasks that require minimal reasoning. 
    To address this, we propose a novel RL framework, \dec, to reduce inefficient reasoning for models. Our method mainly consists of three core components: (1) an innovative advantage decoupled algorithm to guide model reduction of inefficient tokens; (2) a difficulty-aware length penalty to lower the overall length of model responses; (3) an advantage clipping method to prevent bias in policy optimization.
    In our experiments, applied to \distill and \distills as base models, \dec achieves a significant reduction in sequence length by 39\% and reduces excessive reasoning paths in inefficient tokens, while outperforming the base model in overall accuracy. 
\end{abstract}

\section{introduction}

Recent large reasoning models (LRMs) \citep{openai,deepseekr1} have achieved significant advances in mathematical reasoning and programming by leveraging extensive chains of thought (CoT) \citep{cot}. These models emulate human-like deep thinking through mechanisms such as self-reflection, error correction, and the exploration of multiple solution strategies. However, chains of thought often contain long and redundant reasoning trajectories, a phenomenon known as the \Overthink \citep{overthinkdanger}, which leads to substantial inference latency and high computational costs.  
For instance, models may repeatedly verify an already correct answer through redundant self-reflection or unnecessarily complicate simple problems by generating overly elaborate reasoning steps.

To address the overthinking problem, recent approaches can be broadly categorized into three directions. First, some methods construct preference datasets based on output length \citep{DAST} for model training. However, this strategy suffers from preference mismatch and is labor-intensive for data construction. Second, other approaches incorporate a length penalty into the reward function to encourage more concise generation \citep{lead}. While effective in reducing response length, these methods treat the entire model response as a whole and thus fail to guide the model in identifying and suppressing specific redundant reasoning segments. Moreover, the length penalty can distort the advantage estimation of individual tokens, leading policy updates in the wrong direction and ultimately degrading model accuracy. Third, recent work by \citet{lcr1} attempts to mitigate overthinking by extracting valid thinking tokens from the CoT and down-weighting the advantage values of invalid ones. However, this method relies solely on the length ratio between valid and invalid reasoning segments to modulate advantages, without accounting for the specific degree of overthinking in inefficient segments. Consequently, the model struggles to learn how to effectively suppress specific overthinking patterns.

Therefore, building upon the insights and limitations from these works, we propose to partition the model responses into efficient and inefficient segments, enabling us to mitigate overthinking in the inefficient parts while simultaneously reducing overall response length, bypassing the need for labor-consuming preference dataset construction. To achieve this, we propose \textbf{DE}coupled Advantage \textbf{P}olicy \textbf{O}ptimization (\textbf{\dec}), an innovative RL algorithm that introduces three key innovations: (1) we introduce a decoupled advantage computation method for inefficient tokens and down-weight their gradient updates according to the degree of overthinking in the corresponding segment; (2) we implement a difficulty-aware length penalty that encourages shorter responses for easier questions while reducing overall output length; (3) we further propose an advantage clipping strategy to prevent reward fluctuations from steering policy updates in the wrong direction.

\begin{figure}[t]
    \centering
    \includegraphics[width=1\linewidth]{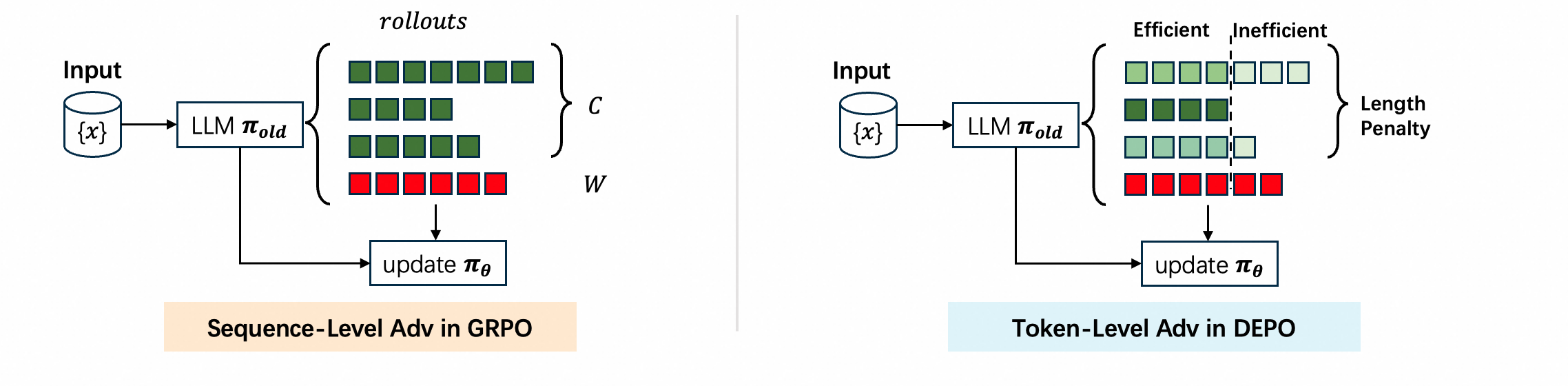}
    \caption{Illustration of proposed \dec. \dec enables token-level advantage estimation to update $\pi_{old}$ to $\pi_{\theta}$, in contrast to sequence-level methods in \grpo.}
    \label{fig:matching}
\end{figure}

Our experiments demonstrate that \dec effectively mitigates overthinking in inefficient reasoning segments, reducing redundant reasoning steps by more than 50\%. Across multiple test sets, \dec consistently shortens model responses while preserving or even slightly improving task accuracy compared to the base model. Specifically, when applied to the \distill and \distills models, \dec achieves an average accuracy gain of 2.0\% over the base model, accompanied by a 38.7\% reduction in response length for \distill and a 39.1\% reduction for \distills. These impressive results indicate that targeting redundant reasoning in inefficient responses can effectively mitigate overthinking in \lrm while preserving training accuracy.

In conclusion, our contributions can be summarized as follows:
\begin{itemize}
    \item We propose a novel algorithm that decouples advantage computation for efficient and inefficient reasoning segments. By leveraging a pretrained GRM, our method precisely identifies the first reasoning step that leads to the correct answer, enabling an explicit separation of reasoning trajectories.
    
    \item We analyze the semantic characteristics of base models during the reasoning process and identify generalizable overthinking patterns across diverse input datasets, thereby enabling a more principled quantification of overthinking tendencies.
    
    \item We introduce an innovative difficulty-aware length penalty and an advantage clipping strategy that jointly prevent distortion in token-level advantage estimation, adaptively reducing response length according to problem difficulty.
\end{itemize}
\section{Preliminary}
Given a prompt $x=[x_1,\ldots,x_n,\textless\text{think}\textgreater]$,
$x=[x_1,\ldots,x_n]$ denotes the user tokens, and \text{\textless think\textgreater} is a special token to trigger the generation of reasoning trajectories \citep{deepseekr1}. \lrm generates a response $y=[y_1,\ldots,y_l,\text{\textless /think\textgreater},y_{l+2},\ldots,y_m]$
where $y=[y_1,\ldots,y_l]$ denotes chains of thought, and $[y_{l+2},\ldots,y_m]$ represents the summary of the long CoT. Typically, the \lrm produces an excessively long CoT, \ie $[y_1,\ldots, y_l]$, which leads to overthinking and unnecessary reasoning trajectories.

We observe that in the naive GRPO framework, the advantage for each rollout is computed as a single sequence-level value, as shown in Eq.\ref{adv}:

\begin{equation}
\label{adv}
    \hat{A}_{i,t}=\hat{A}_i=\frac{r_i-\text{mean(r)}}{\text{std(r)}}
\end{equation}
Here, $\hat{A}_i$ denotes the normalized advantage of the i-th rollout, and all tokens in the response ($[y_1,\ldots,y_m]$) are assigned this identical advantage value, regardless of whether they belong to efficient or inefficient reasoning segments. This design inherently limits the model's ability to distinguish between efficient and inefficient tokens during optimization. To address this limitation, we refine the original response into $y=[y_1,\ldots,y_{ans},y_{ans+1},\ldots,y_l,\text{\textless /think\textgreater},y_{l+2},\ldots,y_m]$, which explicitly separates the reasoning trajectory into efficient and inefficient segments. Specifically, we define $[y_1,\ldots,y_{ans}]$, which first derives the correct answer as an efficient segment. And we define $[y_{ans+1},\ldots,y_l]$, which could contain verification or self-reflection to the correct answer $y_{ans}$ as an inefficient segment.

\section{Methodology}
\label{section_dec}

\begin{algorithm*}[t] 
  \caption{\textbf{DEPO}: \textbf{DE}coupled Advantage \textbf{P}olicy \textbf{O}ptimization for Efficient Thinking} 
  \label{algo}
  \begin{algorithmic}[1] 
    \REQUIRE Initial policy model $\pi_\theta$, generative reward model \grm, task dataset $\mathcal{D}$, hyperparameters $\alpha, \beta$
    
    \FOR{step $= 1, \ldots, M$}
      \STATE Sample a batch $\mathcal{D}_b$ from dataset $\mathcal{D}$ 
      \STATE Sample $G$ outputs $\{o_i\}_{i=1}^G \sim \pi_{\theta_{\text{old}}}(\cdot|x)$ for each prompt $x \in \mathcal{D}_b$ 
      \STATE Compute accuracy reward $R_{\text{accuracy}}$ (Eq.\ref{accuracy}) and length reward $R_{\text{length}}$ (Eq.\ref{length})
      \STATE Compute sequence-level Advantages $\hat{A}_i'$ (Eq.\ref{adv}) and clip biased Advantages $\hat{A}_i$(Eq.\ref{clip})
      \STATE Match redundant reasoning steps in $o_{ie}$ of correct output(Eq.\ref{k}
      ) and compute token-level Advantages $\hat{A}_{i,t}$ (Eq.\ref{token-level})
      \STATE Update the policy $\pi_{\theta}$ by maximum $\mathcal{J}_{\text{DEPO}}(\theta)$ (Eq.\ref{depo})
    \ENDFOR
    
    \ENSURE $\pi_{\theta}$
  \end{algorithmic}
\end{algorithm*}

Our \dec algorithm comprises three key components: (1) an advantage decoupled computation algorithm for efficient and inefficient tokens, which reduces the update weights of inefficient segments; (2) a difficulty-aware length penalty to reduce the overall response length of models; (3) an advantage clipping strategy designed to mitigate gradient bias in policy optimization induced by the length penalty. By employing these methods, \dec effectively identifies and suppresses redundant reasoning, \ie $[y_{ans+1},\ldots,y_{l}]$, and reduces the overall length of the model's responses without misleading gradient update in policy optimization. The overall algorithmic pipeline is outlined in Algorithm~\ref{algo}.
\subsection{Decoupled Advantage}
\begin{wrapfigure}[14]{r}{0.4\textwidth}  
    \centering
    \vspace{-1.2em}  
    \includegraphics[width=\linewidth, height=0.2\textheight]{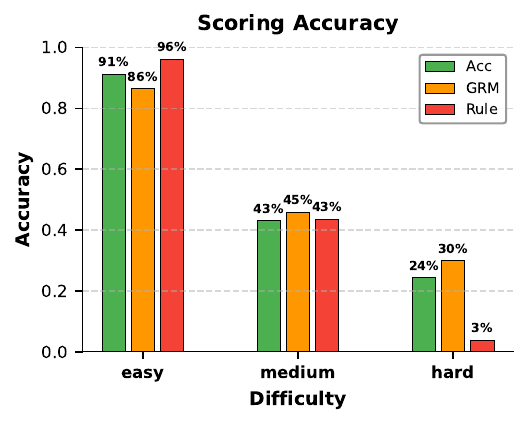}
    \caption{Scoring accuracy of rule and GRM across difficulty levels.}
    \label{fig:scoring}
\end{wrapfigure}
We propose an advantage decoupled computation method that guides the model to learn primarily from efficient tokens while suppressing redundant reasoning. Specifically, we denote the tokens after the sentence, first deriving the correct answer, as the inefficient part, \ie $[y_{ans+1},\ldots,y_l]$. And we fine-tuned a generative reward model (GRM) to accurately identify the token that derives the correct answer ($y_{ans}$) and split off inefficient tokens. Additionally, we substituted GRM for rule to score responses, as GRM significantly outperforms rule-based methods in scoring accuracy. As shown in Fig.\ref{fig:scoring}
, we sampled 1024 rollouts from the DeepScaleR (\citet{luo2025deepscaler}) dataset using \distill and categorized them by difficulty as shown in Fig.\ref{fig:scoring}.

The results show that \grm achieves higher scoring accuracy on challenging tasks with complex answer formats. However, this comes at the cost of increased GPU memory consumption. Detailed usage of \grm is provided in Appendix.B.\par

Additionally, we reduce the advantage value of inefficient tokens based on the number of redundant reasoning paths they contain. To this end, we analyze responses from \distill to catalog patterns of overthinking behavior, and develop a quantifiable method for identifying overthinking and redundant reasoning steps within the segment $[y_{ans+1},\ldots,y_{l}]$, as illustrated in Fig.\ref{fig:matching}, in which $N$ denotes the maximum of transition phrases that start an alternative reasoning step, and $X$ denotes the total number of self-reflection words in $o_{ie}$.

And the redundant reasoning steps in $o_{ie}$ is formulated as:
\begin{equation}
\label{k}
    K=\text{max}(N,X)
\end{equation}

For a given prompt $x$ and generated output $o_i$, the final loss functino of \dec is:

\begin{equation}
\label{depo}
\begin{aligned}[b]
\mathcal{J}_{\text{DEPO}}(\theta) = \mathbb{E}_{x \sim D, \{o_i\}_{i=1}^G \sim \pi_{\theta_{\text{old}}}(\cdot|x)} &\bigg[ \frac{1}{\sum_{i=1}^G |o_i'|} \sum_{i=1}^G \sum_{t=1}^{|o_i'|} \\
&\Bigg\{ \min\bigg[ \frac{\pi_{\theta}}{\pi_{\text{old}}} \cdot \hat{A}_{i,t},\, \text{clip}(\frac{\pi_{\theta}}{\pi_{\text{old}}}, 1\!-\!\epsilon, 1\!+\!\epsilon) \cdot \hat{A}_{i,t} \bigg] 
 \Bigg\} \bigg]
\end{aligned}
\end{equation}

\begin{figure}[t]
    \centering
    \includegraphics[width=1\linewidth]{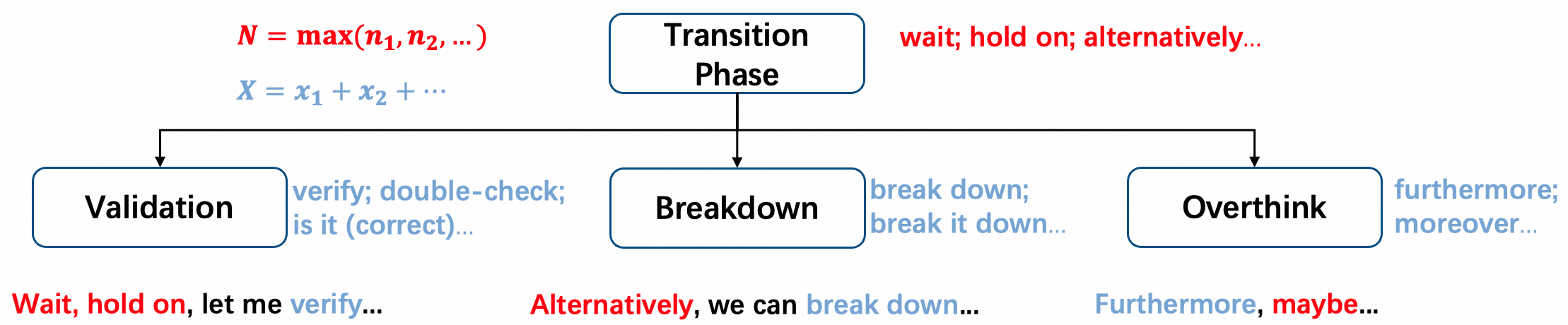}
    \caption{The redundant reasoning matching method of \dec}
    \label{fig:matching}
\end{figure}

We convert the original sequence-level advantage value ($\hat{A}_i$) to a token-level advantage value ($\hat{A}_{i,t}$) compared to the objective of naive \grpo. Specifically, we decompose the reasoning process into distinct segments: an efficient segment $o_{e}=[y_1.\ldots,y_{ans}]$, which directly leads to the correct answer, and an inefficient segment $o_{ie}=[y_{ans+1},\ldots,\textless\text{/think}\textgreater]$, representing overthinking steps. To decouple the contribution of these segments, we lower the advantage values of $o_{ie}$ based on the number of its redundant reasoning steps. The advantage computation method of \dec is defined as:
\begin{equation}
\label{token-level}
    \hat{A}_{i,t}=
\begin{cases}
    f(o_{ie})\cdot\hat{A}_i,  &\text{if }y_t\text{ in }o_{ie}\text{ and }o_i\text{ is correct}\\
    \hat{A}_i,  &\text{otherwise}
\end{cases}
\end{equation}
where $\hat{A}_i$ is the sequence-level advantage of response $o_i$ and $f$ is formulated as:

\begin{equation}
\label{f}
    f(\cdot)=1-\beta\cdot(1-e^{-\beta\cdot K})
\end{equation}

where $\beta$ is hyperparameter and $K$ denotes the number of redundant reasoning steps identified in $o_{ie}$ as presented in Eq.\ref{k}, through our predefined rule-based method.
The range of function $f(\cdot)$ is dependent on $\beta$, and $f(\cdot)$ decreases monotonically with the increase of $K$. 

Following the experimental setup of DAPO (\citet{dapo}), we compute the final loss across all tokens within a group, while explicitly excluding the KL Divergence term to better enhance models' reasoning capabilities.

\subsection{Difficulty-Aware Length Penalty}
In addition to the aforementioned method for reducing redundant reasoning, we also incorporate a length penalty mechanism to minimize the overall response length of the model. The core idea is to reward shorter responses and penalize longer ones within a group of rollouts, particularly for simple questions that require only a minimal number of reasoning tokens. Given a prompt $x$ and a group of rollouts $\{o_1, \ldots, o_G\}$, we denote their respective lengths as $\{l_1, \ldots, l_G\}$. The accuracy reward is then defined as follows:
\begin{equation}
\label{accuracy}
   R_{\text{accuracy}}(o_i \mid x) = 
\begin{cases} 
1, & \text{if } o_i \text{ is correct} \\
0,                & \text{if } o_i \text{ is incorrect} \\
-1, &\text{if } o_i \text{ is overlong}
\end{cases} 
\end{equation}
We introduce a negative reward $R_{\text{accuracy}} = -1$ for responses that exceed the maximum allowed response length, treating such cases as more severe failures than merely incorrect answers. This design is motivated by our preliminary experiments, which revealed that approximately 10\% of model generations exhibited excessive repetition, resulting in abnormally long outputs. \par
\myparagraph{Length Penalty} Furthermore, we introduce a length reward based on the length variance among correct responses and the difficulty of question $x$, and we present the formulation as follows:
\begin{equation}
\label{length}
   R_{\text{length}}(o_i \mid x) = 
\begin{cases} 
-\alpha\cdot(1-e^{-\alpha\cdot \delta})\cdot\frac{\mid o_i\mid-\text{mean(}l_{\text{pos}}\text{)}}{\text{std(}l_{\text{pos}}\text{)}}, & \text{if } o_i \text{ is correct} \\
0,                & \text{if } o_i \text{ is incorrect} 
\end{cases} 
\end{equation}
where $\alpha$ is a hyperparameter, $l_{pos}$ denotes the lengths of correct responses, and $\delta$ represents the number of correct responses in a group of rollouts, which means the difficulty of input question $x$. So the final reward of response $o_i$ is:

\begin{equation}
    R(o_i\mid x)=R_{\text{accuracy}}(o_i\mid x)+R_{\text{length}}(o_i\mid x)
\end{equation}\par

\subsection{Advantage Clipping} 
\label{method:adv_clip}
\noindent
\begin{minipage}[]{0.58\textwidth}
    In the computation of advantages compared to naive \grpo, we introduce a clipping operation to the original computation method. As revealed in our preliminary experiments in Fig.\ref{adv_clip}, length-based reward signals, \eg $R_{accuary}=-1$ for excessively long responses and $R_{length}$ for correct responses, could introduce advantage estimation biases: (1) correct responses which are penalized for excessive length may yield negative advantages; (2) incorrect responses could exhibit positive advantages when co-occurring with responses that exceed the maximum length. These cases could mislead policy updates, ultimately causing accuracy degradation. To address this, we propose the following clipping method:
\end{minipage}
\hfill
\begin{minipage}[]{0.4\textwidth}
    \centering
    \includegraphics[width=\linewidth]{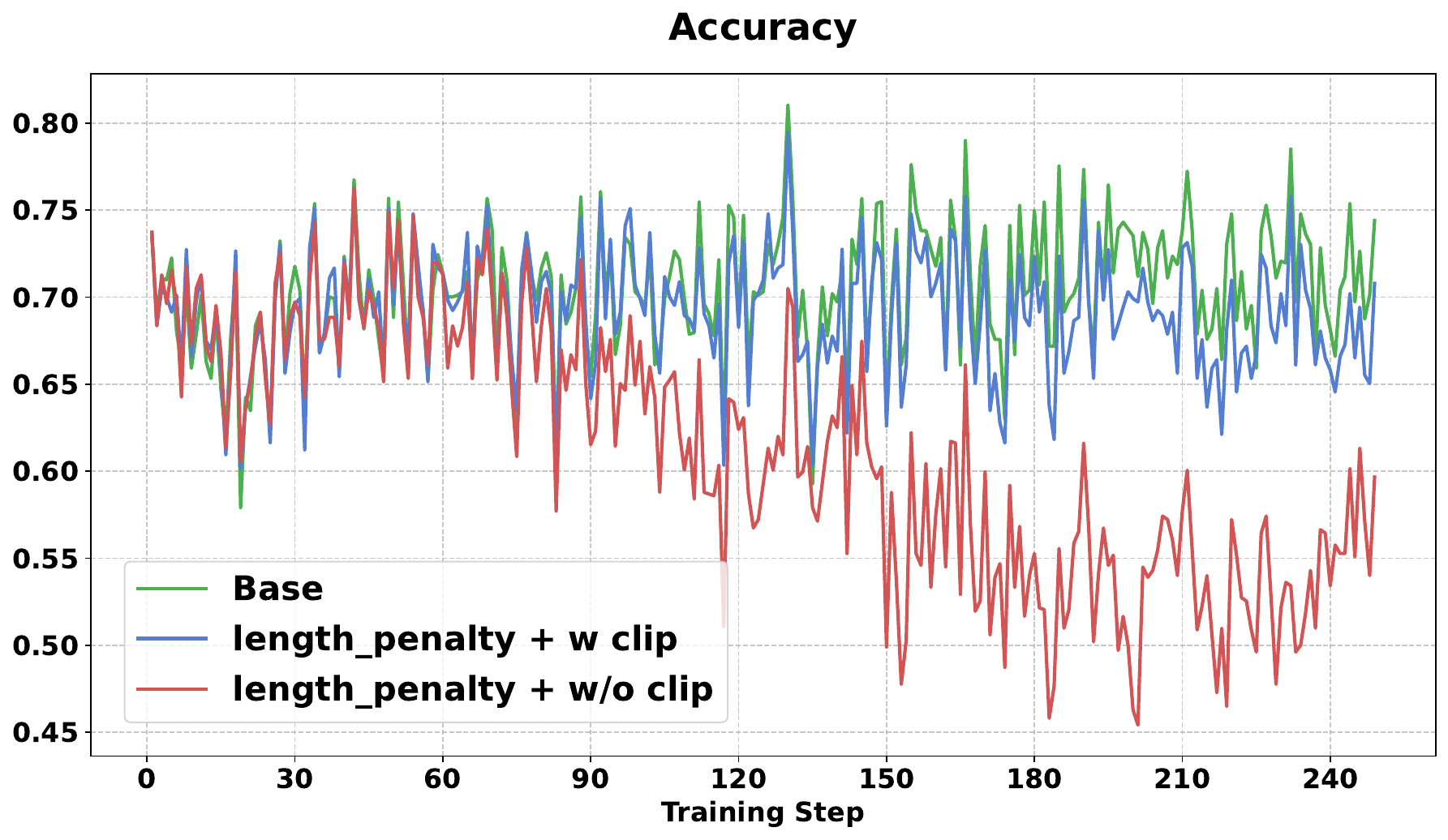}
    \captionof{figure}{Training accuracy comparison of naive GRPO, length penalty with \textit{adv\_clip} and length penalty without \textit{adv\_clip}.}
    \label{adv_clip}
\end{minipage}

\begin{equation}
\label{clip}
    \hat{A}_i=
\begin{cases}
    \text{clip}(\hat{A}_i',\text{min}(\hat{A}_{\text{pos}}'),+\infty),   &\text{if }o_i \text{ is correct} \\
    \text{clip}(\hat{A}_i',-\infty,0), &\text{if } o_i \text{ is incorrect}
\end{cases}
\end{equation}
where $\hat{A}_i'$ is the original advantage of response $o_i$, and $\text{min}(\hat{A}_{pos}')$ denotes the minimum positive value among unclipped advantage values in $\{o_1,\ldots,o_G\}$. Our method ensures that correct answers consistently yield positive advantages, whereas incorrect answers produce strictly negative values, thereby preventing the model's gradient updates from being trapped in conflicting optimization directions.

\section{Experiments}
\subsection{SetUp}
\myparagraph{Model}We adopt \distill (\citet{deepseekr1}) and \distills (\citet{deepseekr1}) as our policy model, which has superior performance in mathematical problems and exhibits twice the length of reasoning compared to its base model. 

\myparagraph{Dataset and Metric}We select DeepScaleR (\citet{luo2025deepscaler}) as our training dataset, which consists of approximately 40,000 unique mathematics problem-answer pairs compiled from AIME 1984-2023, AMC (prior to 2023), Omni-Math (\citet{omni}), and STILL (\citet{still}). For the evaluation task, we adopt four math datasets as our test datasets: AMC23, MATH500 (\citet{math500}), AIME24 and AIME25. Given the limited size of samples in AIME24, AIME25, and AMC23 (30, 30, and 40 instances, respectively), we repeatedly sample each case in these datasets 16 times and adopt the average accuracy (avg@16) as the evaluation metric. For the remaining datasets, \ie MATH500, we uniformly used pass@1 as the metric. The top-p and temperature of the evaluation task are 0.95 and 0.7, and the maximum context size is 16K.

\myparagraph{\grm}To accurately score the model's responses and extract the first reasoning sentence that leads to the correct answer, we fine-tuned the Qwen2.5-Instruct-7B model (\citet{qwen2025qwen25technicalreport}) via Supervised Fine-Tuning (SFT) to serve as the \grm. Detailed fine-tuning procedures are provided in Appendix.C.

\myparagraph{Implementation}We conducted experiments on the VeRL framework (\citet{verl}). The maximum response length, training batch size and learning rate are set to 16K, 128, 1e-6 respectively, and the hyperparameters of $\alpha$, $\beta$, and the number of rollouts $G$ are 0.2, 0.5 and 8, respectively. We implement alternating execution of the GRM scoring and policy model training tasks through vLLM offload, with the \grm's top-p and temperature parameters set to 0.95 and 1, respectively. The comparative analysis of length penalty versus advantage decoupled computation is detailed in the ablation experiment. Additionally, we use one 8$\times$H20 node to train the model for 1 epoch, and we select the checkpoint that achieves an optimal balance between response length and accuracy during training as the baseline for comparison.
\subsection{Baselines}

We compared the performance of \dec with the following methods in terms of accuracy and model response length:
\begin{itemize}
    \item \textbf{GRPO} (\citet{grpo}) proposes a group-related optimization algorithm that computes sequence-level advantages from a group of rollouts for policy optimization.
    \item \textbf{DAST} (\citet{DAST}) constructs preference data by ranking pre-sampled responses using a length-based reward function, and then applies SimPO (\citet{simpo}) to fine-tune the model. 
    \item \textbf{GRPO-LEAD} (\citet{lead}) introduces a length-dependent accuracy penalty to promote concise generation and an explicit penalty mechanism for incorrect responses, and re-weights advantage values based on problem difficulty, and then fine-tunes the model via GRPO.
    \item \textbf{LC-R1} (\citet{lcr1}) first calculates a length reward based on the ratio of the response length to the maximum length within the same group. Then, it extracts valid thinking tokens from the CoT process using the LC-Extractor module, which are treated as the compression part. Finally, it trains the model by computing the loss on the compression and invalid thinking part, respectively.
\end{itemize}
\subsection{Experiment Results}
\begin{table}[h]
  \centering
  \caption{Accuracy (Acc) and response length (Length) of different methods on AIME24, AIME25, AMC23, MATH500. The best and second results are bold and underlined respectively.}
  \label{tab:method_performance}
  \renewcommand{\arraystretch}{1.3}
  \setlength{\tabcolsep}{3pt}

  \newcolumntype{Y}{>{\centering\arraybackslash}X}

  \begin{tabularx}{\linewidth}{@{} l | Y Y | Y Y | Y Y | Y Y | Y Y @{}}
    \toprule
    \multirow{2}{*}{Method} 
      & \multicolumn{2}{c|}{AIME24} 
      & \multicolumn{2}{c|}{AIME25} 
      & \multicolumn{2}{c|}{AMC23} 
      & \multicolumn{2}{c|}{MATH500} 
      & \multicolumn{2}{c}{Avg} \\
    \cmidrule(lr){2-3} \cmidrule(lr){4-5} \cmidrule(lr){6-7} \cmidrule(lr){8-9} \cmidrule(lr){10-11}
    & Acc & Length & Acc & Length & Acc & Length & Acc & Length & Acc & Length \\
    \midrule
    \multicolumn{11}{@{}l@{}}{DeepSeek-R1-Distill-Qwen-7B} \\
    \midrule
    Original & 49.6 & 10670 & \underline{39.8} & 11068 & 87.8 & 5794 & 92.8 & 3601 & 69.3 & 7591 \\
    GRPO & \textbf{55.0} & 9913 & \textbf{42.0} & 10709 & 88.1 & 5613 & 93.4 & 3527 & \textbf{71.3} & 7263 \\
    DAST & \underline{52.9} & 9674 & 36.0 & 10729 & 88.8 & 5091 & 93.6 & 2904 & 69.7 & 6906 \\
    LC\_R1 & 49.2 & \underline{7013} & 37.3 & \underline{7530} & 87.3 & \textbf{2847} & 93.4 & \textbf{2276} & 68.6 & \underline{4733} \\
    GRPO-Lead & 49.2 & 9507 & 39.2 & 9529 & \underline{89.0} & 4525 & \underline{93.8} & 2957 & 69.7 & 6434 \\
    DEPO & 52.7 & \textbf{6580} & 39.2 & \textbf{7092} & \textbf{90.5} & \underline{3215} & \textbf{94.4} & \underline{2318} & \underline{71.1} & \textbf{4656} \\

    \midrule
    \multicolumn{11}{@{}l@{}}{DeepSeek-R1-Distill-Qwen-1.5B} \\
    \midrule
    Original & 30.2 & 12165 & 24.4 & 12109 & 70.0 & 7720 & 84.6 & 4820 & 54.0 & 9048 \\
    GRPO & \textbf{33.5} & 10609 & \textbf{27.3} &10668 &\textbf{74.4} & 6561 & 86.8 & 4208 & \textbf{57.2} & 7864 \\
    DEPO & 30.8 & \textbf{7732} & 24.8 & \textbf{7649} & 74.2 & \textbf{4388} & \textbf{87.2} & \textbf{2762} & 56.1 & \textbf{5510} \\

    \bottomrule
  \end{tabularx}
\end{table}
In this section, we present a comprehensive evaluation comparing \dec with baseline methods across multiple dimensions. As shown in Table.\ref{tab:method_performance}, for \distill, \dec achieves substantial length reduction of 38.3\% and 35.9\% on challenging problem sets, \ie AIME24 and AIME25, attaining the shortest generated length among all compared baselines. Besides, regarding the accuracy metric, \dec shows stable performance compared to the original model. With accuracy fluctuations ranging from -0.6\% regression and +3.1\% improvement across challenging datasets. However, \dec incurs approximately 3\% loss of average accuracy versus naive GRPO on challenging datasets, as its aggressive length optimization may compromise complex reasoning steps required for these tasks. Furthermore, in simple datasets (AMC23, MATH500), \dec achieves the highest accuracy while maintaining second-best length efficiency, demonstrating optimal performance for routine tasks requiring both precision and conciseness. For the smaller \distills model, \dec reduces response length by 39.1\% while achieving an accuracy that is only 0.2\% lower than \grpo and 2.1\% higher than the base model, effectively improving accuracy while substantially shortening model responses.

\begin{figure}[h]
    \centering
    \includegraphics[width=0.98\linewidth]{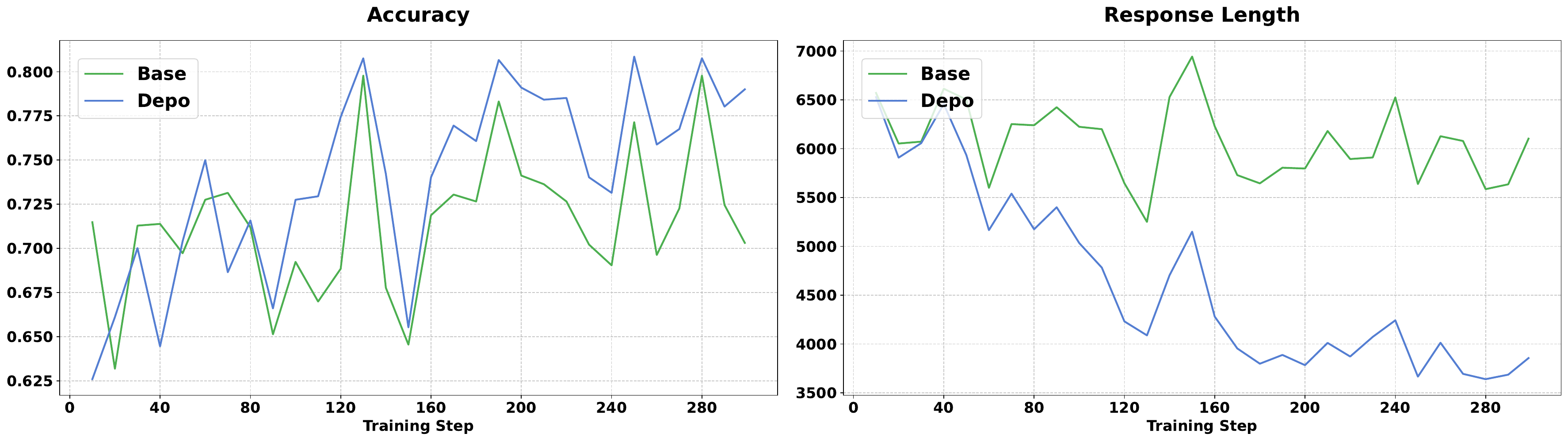}
    \caption{Comparison of model accuracy and response length on DeepScaleR between naive \grpo and \dec at different training steps.}
    \label{acc_len}
\end{figure}
During the training process on \distill as shown in Fig.\ref{acc_len}, although the scoring accuracy of \grm and the rule-based method shows minor discrepancy (3-5\% divergence), their overall trends of scoring remain aligned. Besides, \dec significantly reduces model response length during training, decreasing the average from approximately 6,500 tokens to around 3,600 tokens, which represents a 44\% reduction in sequence length.

\begin{figure}[h]
    \centering
    \includegraphics[width=0.98\linewidth]{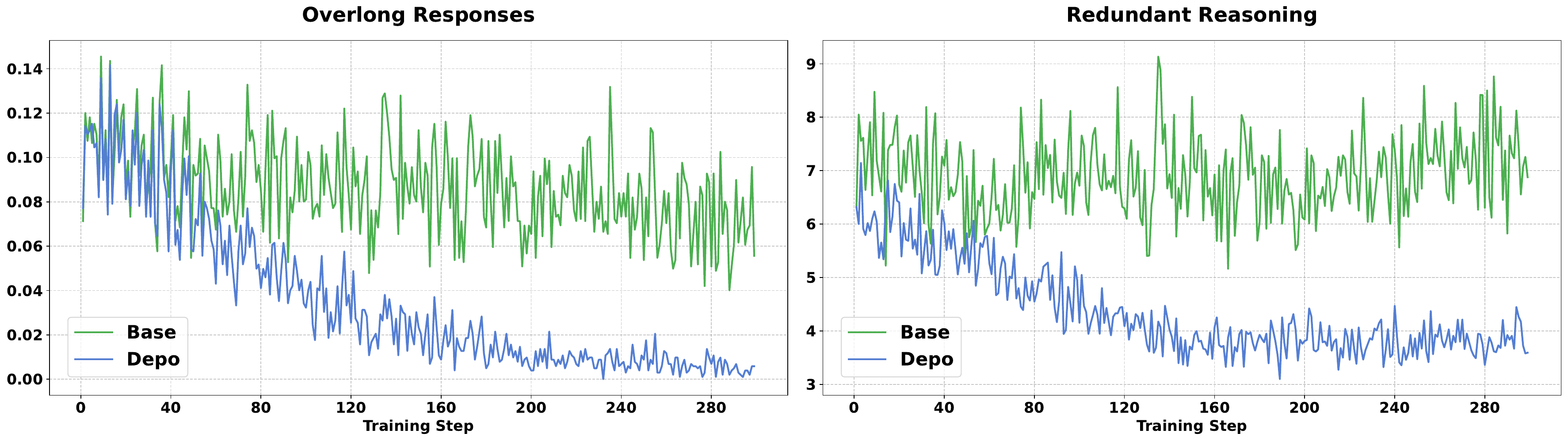}
    \caption{Comparison of overlong responses ratios and redundant reasoning steps in inefficient segments per rollout sample between naive \grpo and \dec.}
    \label{overlong_reason}
\end{figure}

\myparagraph{\dec suppresses repetitive outputs and redundant self-reflection compared to naive \grpo}As shown in Fig.\ref{overlong_reason}, our experimental results demonstrate that \dec effectively mitigates repetition-induced overlong responses during generation. Compared to baseline GRPO training—which produces approximately 110 overlong outputs per 1024 rollout (accounting for  10.7\% of total generations), \dec reduces this frequency to just 1-2 occurrences (0.1\% of rollouts), representing a 98\% decrease in overlong outputs. 
The phenomenon observed in these cases primarily arise from two key factors: first, repetitive verification—where the model’s self-reflection behavior leads to redundant validation in its reasoning process (e.g., repeated phrases like “let’s check again” or “wait, hold on”), often resulting in excessively long responses; second, auto regressive error accumulation in \lrm, where gradual error buildup during generation causes the probability of certain tokens to incrementally increase, leading to unintended repetition of those tokens in the sequence. \dec effectively mitigates this issue by imposing a stricter penalty on repetitive patterns, directly addressing both the reflective loops from self-reflection and the token-level redundancy from auto regressive errors.

\myparagraph{\dec Reduces Redundant Reasoning in Inefficient Segments}As shown in the right panel of Fig.\ref{overlong_reason}, \dec significantly reduces redundant reasoning steps—where the model keeps re-verifying a correct answer—by using a rule-based matching mechanism (illustrated in Fig.\ref{fig:matching}). This approach reduces redundant verification by about half compared to \grpo, significantly improving the efficiency of the model’s reasoning process. 

\subsection{Ablation Results}
\label{ablation section}
We evaluate the contribution of each component of \dec, \ie Adv\_Decouple and Len\_Penalty, through an ablation experiment on \distill. We present the ablation results in Table.\ref{tab:ablation_study}
\begin{table}[h]
  \centering
  \caption{Ablation Study of Advantage Decoupling (Adv-Decouple) and Length Penalty (Len-Penalty) on Model Accuracy and Redundant Reasoning}
  \label{tab:ablation_study}
  \renewcommand{\arraystretch}{2.5}
  \setlength{\tabcolsep}{3.4pt}
  \begin{tabular}{l | rr | rr | rr | rr | rr}
    \toprule
    \multirow{2}{*}{Method} & \multicolumn{2}{c|}{AIME24} & \multicolumn{2}{c|}{AIME25} & \multicolumn{2}{c|}{AMC23} & \multicolumn{2}{c|}{MATH500} & \multicolumn{2}{c}{Avg} \\
    \cmidrule(lr){2-3} \cmidrule(lr){4-5} \cmidrule(lr){6-7} \cmidrule(lr){8-9} \cmidrule(lr){10-11} 
    & \makecell{Acc \\Len} & Reflect & \makecell{Acc \\Len} & Reflect & \makecell{Acc \\Len} & Reflect & \makecell{Acc \\Len} & Reflect  & \makecell{Acc \\Len} & Reflect \\
    \midrule 
    \dec               & \makecell{\textbf{52.7}\\\textbf{6580}} & 5.2 & \makecell{\textbf{39.2}\\\textbf{7092}} & 5.5 &\makecell{\textbf{90.5}\\\textbf{3215}} &\textbf{2.1} & \makecell{94.4\\\textbf{2318}} &\textbf{1.6}&  \makecell{\textbf{71.1}\\\textbf{4656}} & 3.5 \\
    \midrule 
    -w/o Adv-Decouple   & \makecell{50.4\\7002}  &6.8             & \makecell{37.7\\7300}        &7.3       & \makecell{86.6\\3450}&2.8  & \makecell{93.8\\2620}  &3.1         &   \makecell{68.9\\4944} & 4.8\\
    \midrule 
    -w/o Len-Penalty    & \makecell{52.1\\6962} &\textbf{5.1}              & \makecell{38.8\\7638}  &\textbf{5.2}             & \makecell{88.4\\3721}   &\textbf{2.1}         & \makecell{\textbf{94.8}\\2953} &1.7&   \makecell{70.3\\5174} & \textbf{3.4} \\
    \bottomrule 
  \end{tabular}
\end{table}

\myparagraph{The accuracy and length trade-off of length penalty}As shown in Table.\ref{tab:ablation_study}, the ablation variant without advantage decoupling (w/o Adv-Decouple)—which relies solely on length penalty—produces consistently shorter responses than the variant without length penalty (w/o Len-Penalty). This indicates that the length penalty is more effective at reducing output length than our advantage decoupling mechanism alone, yielding an average additional reduction of about 300 tokens across most datasets. However, this comes at a slight cost in accuracy: even with advantage clipping (Sec.\ref{method:adv_clip}), the length-penalty-only model underperforms the advantage decoupling model. These results highlight the need to carefully balance response length and accuracy when applying length penalties.

\myparagraph{Advantage decoupling versus length penalty in suppressing self-reflection}We further quantify the number of redundant reasoning steps (e.g., “double-check” or “wait, hold on”) in the inefficient reasoning segments ($o_{ie}$) across ablation settings. The results show that advantage decoupling reduces such redundant behaviors more effectively than the length penalty, leading to fewer unnecessary verification steps or shifts to alternative reasoning paths after the model has already reached the correct answer. Moreover, in the full \dec model, advantage decoupling remains the dominant component for both suppressing self-reflection and improving accuracy—outperforming length penalty in these aspects.
\section{Related Work}
\myparagraph{Reasoning of CoT in \lrm}Following \citet{cot}'s demonstration that extended Chain-of-Thought (CoT) enhances Large Reasoning Models (LRMs), frontier models like OpenAI o1 (\citet{openai}) and DeepSeek-R1 (\citet{deepseekr1}) now employ reinforcement learning to fine-tune reasoning trajectories. During this process, models dynamically verify and switch reasoning paths—termed \textit{aha-moment} (\citet{deepseekr1})—when encountering solution uncertainty. However, persistent \textit{aha-moments} after correct answers cause excessive verification of already-correct solutions and lead to redundant, lengthy responses, particularly detrimental in mathematical and coding benchmarks, which we denote as \textit{overthink} (\citet{overthinkdanger}).

\myparagraph{Efficient Reasoning for \lrm}Recent methods for improving reasoning efficiency and mitigating overthinking in \lrm typically aim to reduce output tokens. This is achieved through reward shaping based on response length, either by rewarding shorter rollouts during training (\citet{arora}) or setting a "best-length" threshold (\citet{learntoreason}). Alternatively, other approaches include fine-tuning on length preference pairs (\citet{DAST})or employing prompt engineering to elicit shorter responses (\citet{prompting}). More novel techniques encourage models to dynamically decide whether to use Chain-of-Thought (CoT) based on problem difficulty (\citet{adapthink}), or extract only valid reasoning segments for training via auxiliary modules (\citet{lcr1}). However, these works mentioned treat model output as a whole during training, failing to distinguish between efficient and inefficient reasoning segments or evaluate the precise self-reflection mechanisms of overthinking. Motivated by this, we propose \dec in our work, which reduces output length and redundant self-reflection by decoupling their advantage computations for efficient and inefficient reasoning components according to the degree of overthinking, which is a novel direction for efficient reasoning.
\section{Limitation}
In this section, we discuss several limitations of our work: (1) Our training is confined to mathematical datasets since they are easy to verify, and the model's responses contain explicit reasoning steps that derive the correct answer. Further studies are needed to evaluate the effectiveness of \dec on other domains such as logical and code problems. (2) Due to limitations in computational resources and the variance of overthinking between different models, we only conducted experiments on \distill and \distills. Nevertheless, \dec still demonstrated its efficacy for reducing response length and redundant reflection in reasoning steps. Besides, since \dec relies on \grm for both scoring and identifying the first correct reasoning step; thus, its performance critically depends on \grm quality. We address this by rigorously filtering the \grm training data, with details provided in Appendix.C.

\section{Conclusion}
In this paper, we propose \dec, a reinforcement learning algorithm designed to mitigate overthinking in \lrm. \dec addresses this issue through two core mechanisms: (1) it decouples the loss computation for inefficient reasoning segments from the rest of the response, enabling the model to explicitly learn to suppress redundant tokens; (2) it incorporates a length penalty into the reward function, encouraging the model to generate shorter outputs. Extensive experiments demonstrate that \dec effectively balances accuracy and response length, while significantly reducing redundant reasoning steps across diverse mathematical problem sets. 

\bibliography{iclr2026_conference}
\bibliographystyle{iclr2026_conference}
\clearpage
\appendix
\section{Detailed Usage of Generative Reward Model (\grm)}
\label{app:grm}

The Generative Reward Model (GRM) is a core component of \dec, designed to evaluate the quality of model responses and identify efficient vs. inefficient reasoning segments. The detailed usage and prompt of \grm is as follows:
\begin{figure}[h]
    \centering
    \includegraphics[width=0.98\linewidth]{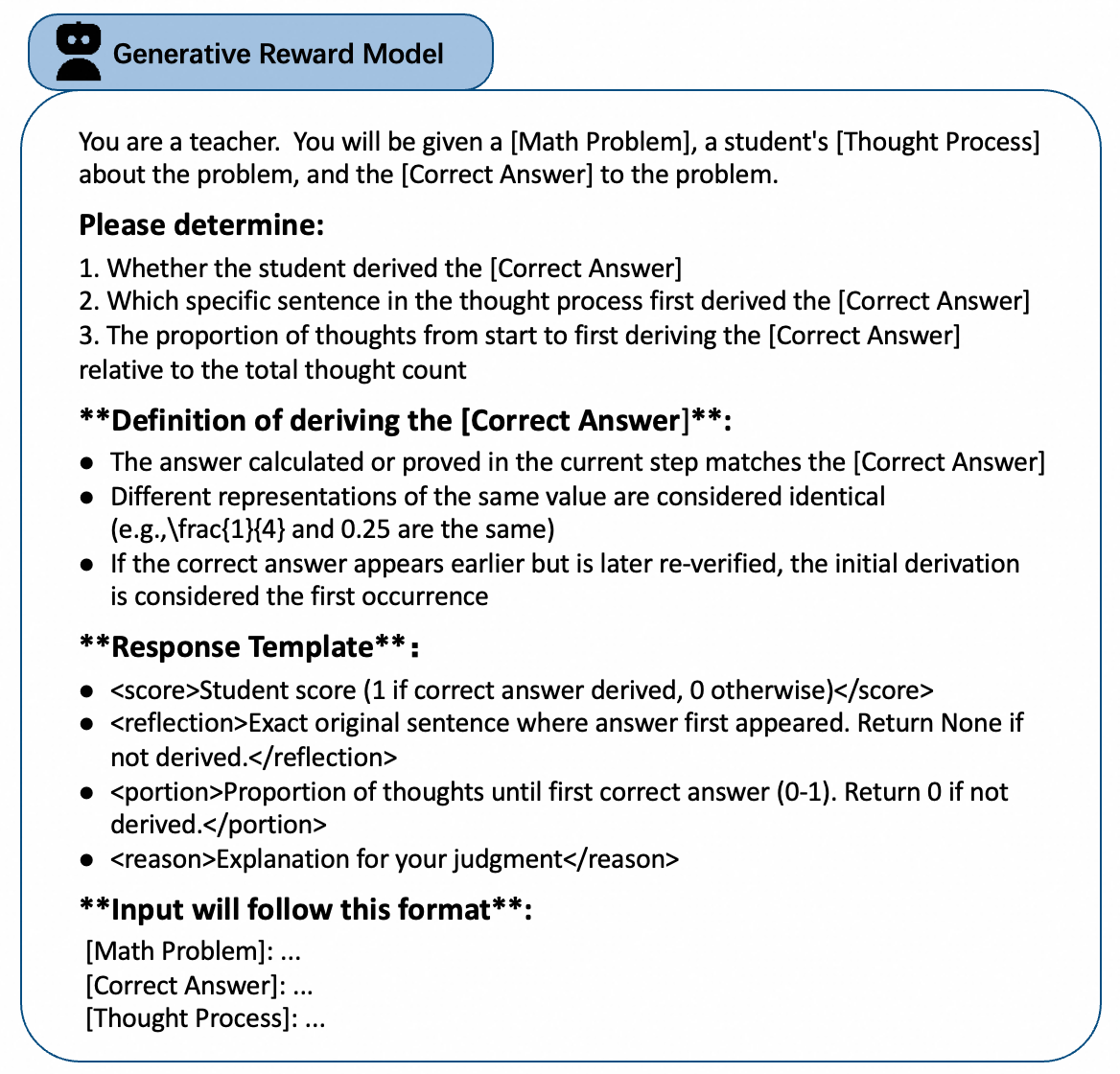}
    \caption{The detailed usage and prompt of \grm.}
    \label{fig:grmusage}
\end{figure}

As shown in Fig. \ref{fig:grmusage}, we provide \grm with a mathematical problem and its corresponding answers, along with the reasoning process generated by \lrm, \ie Chain-of-Thought (CoT). And we have determined the criteria for identifying the initial reasoning step that arrives at the correct answer, requiring \grm to output the following responses:
\begin{itemize}
    \item \textbf{Score}: Score represents the \grm 's assessment of the reasoning correctness of CoT, where a value of 1 indicates that \lrm arrived at the correct answer, and 0 otherwise.
    \item \textbf{Reflection}: Reflection represents the first sentence in CoT that derives the correct answer, which is the distinguishing criterion of efficient and inefficient parts.
    \item \textbf{Portion}: Portion denotes \grm 's estimated ratio of efficient reasoning to the entire length of CoT, providing a fallback mechanism in case the exact "Reflection" matching is unavailable.
    \item \textbf{Reason}: Reason constitutes the \grm 's explanation for its output, enabling us to verify the accuracy of "Score" and "Reflection".
\end{itemize}

\section{Training and Evaluation of \grm}
\label{app:grmtraining}
\subsection{Base Model of \grm}
To accurately score the \lrm's responses and extract the first reasoning sentence leading to the correct answer, we employed Qwen2.5-Instruct-7B as the base model for \grm and conducted Supervised Fine-Tuning using a high-quality dataset, ensuring \grm adheres to our specified response format while enhancing its evaluation accuracy in both scoring and reasoning sentence matching.
\subsection{Dataset and Evaluation of \grm}
To generate a high-quality dataset, we first leveraged \distill to generate 39,961 mathematical problem-response pairs from the DeepScaleR dataset. And we used Qwen2.5-72B model to produce corresponding responses according to the specified format in Fig. \ref{fig:grmusage}, generating score, reflection, portion and reason fields for all pairs. To enhance dataset quality and ensure Qwen2.5-Instruct-7B strictly adheres to our format while improving its scoring and matching accuracy, we implemented rigorous filtering by removing: (1) samples with incorrect scores, (2) responses failing to identify the initial correct reasoning step in CoT, (3) sequences where the portion values deviated by over 0.15 from ground-truth effective ratios, ultimately retaining 18,416 high-quality samples for Supervised Fine-Tuning to derive \grm. Furthermore, post evaluation on the Math500 dataset revealed that \grm correctly scored 474 accurately answered samples among 500 total responses of \distill, successfully matched the first correct reasoning sentence in CoT for 445 samples, achieving 93.9\% matching rate, and maintained portion deviations within 0.15 of ground-truth ratio for 80\% of cases.

In our training process of \lrm, we set the temperature and topp of \grm to 1.0 and 0.95, respectively, and we set the context size of the prompt to 16K and the size of maximum to 1K, since the CoT of \distill might be lengthy and the output formats of \grm are specified and concise.

\section{Case Study}
\begin{figure}[h]
    \centering
    \includegraphics[width=0.98\linewidth]{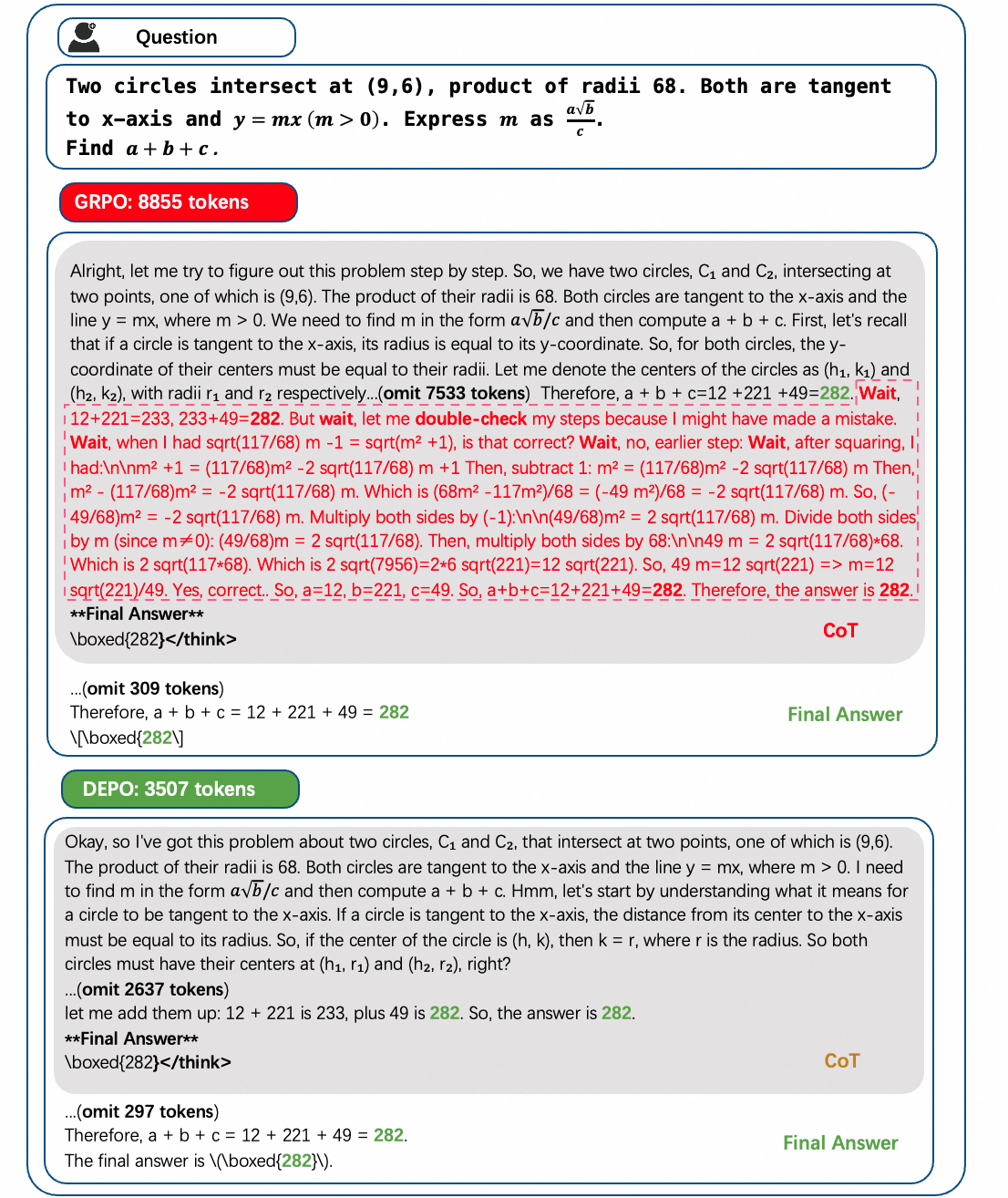}
    \caption{Case study of the comparison of \dec and naive \grpo.}
    \label{fig:case}
\end{figure}

We make a case study in Fig. \ref{fig:case} to compare \dec and the naive \grpo. As illustrated in Fig. \ref{fig:case}, \dec requires fewer tokens than \grpo to derive the correct answer, and \dec can immediately halt the thinking process in CoT while \grpo continues to perform a reflection and verification even after the model has derived the correct answer.

\end{document}